%% file: root.tex
\documentclass[letterpaper, 10pt, conference]{ieeeconf}

\IEEEoverridecommandlockouts
\usepackage{amsmath}
\usepackage{amssymb}
\usepackage{graphicx}
\usepackage{svg}
\usepackage{booktabs}        %
\usepackage{caption}
\usepackage{colortbl}        
\usepackage{xcolor}

\usepackage[table]{xcolor}

\usepackage{booktabs}      %
\usepackage{graphicx}      %
\usepackage{amsmath}       %

\expandafter\def\expandafter\normalsize\expandafter{%
    \normalsize%
    \setlength\abovedisplayskip{2pt}%
    \setlength\belowdisplayskip{2pt}%
    \setlength\abovedisplayshortskip{2pt}%
    \setlength\belowdisplayshortskip{2pt}%
}

\definecolor{TechTeal}{HTML}{E0F2F1}  %
\definecolor{StdGray}{HTML}{757575}   %
\definecolor{DropRed}{HTML}{E53935}   %
\definecolor{RiseGreen}{HTML}{43A047} %
\colorlet{SimGray}{gray!15}           %

\newcommand{\res}[2]{#1\,\textcolor{StdGray}{\scriptsize{(#2)}}}
\newcommand{\cmark}{\textcolor{StdGray}{$\checkmark$}}
\newcommand{\best}[2]{\textbf{#1}\,\textcolor{StdGray}{\scriptsize{(\textbf{#2})}}}

\newcommand{\avg}[1]{#1}
\newcommand{\bestavg}[1]{\textbf{#1}}

\newcommand{\na}{\textcolor{StdGray}{---}} 

\newcommand{\cg}{\cellcolor{SimGray}} 

\newcommand{\drop}[1]{\textcolor{DropRed}{\textbf{\scriptsize{\,$\downarrow$#1}}}}

\newcommand{\rise}[1]{\textcolor{RiseGreen}{\textbf{\scriptsize{\,$\uparrow$#1}}}}

\usepackage{hyperref}
\hypersetup{hidelinks}

\title{\LARGE \bf
GAP: Geometric Anchor Pre-training for\\Data-Efficient Visuomotor Learning of Manipulation Tasks}

    \author{Davide Buoso, Andrea Protopapa, Stefano Di Carlo, Francesca Pistilli, and Giuseppe Averta \\ \textit{Department of Control and Computer Engineering, Polytechnic University of Turin, Italy}
    \thanks{Corresponding author: Davide Buoso (e-mail: davide.buoso@polito.it).}
\thanks{This work was carried out within the Future Artificial Intelligence Research (FAIR) and received funding from the European Union Next-GenerationEU (PIANO NAZIONALE DI RIPRESA E RESILIENZA (PNRR) – MISSIONE 4 COMPONENTE 2, INVESTIMENTO 1.3 – D.D. 1555 11/10/2022, PE00000013. This manuscript reflects only the authors’ views and opinions, neither the European Union nor the European Commission can be considered responsible for them. We acknowledge the CINECA award under the ISCRA initiative, for the availability of high performance computing resources and support.}}

\renewcommand{\baselinestretch}{0.983}

\begin{document}

\maketitle
\thispagestyle{empty}
\pagestyle{empty}

\input{sections/0_abstract}

\input{sections/1_intro}
\input{sections/2_related}
\input{sections/3_method}
\input{sections/4_exps}
\input{sections/5_conclusions}

\addtolength{\textheight}{-2cm}  %

\bibliographystyle{IEEEtran}
\bibliography{references}

\end{document}

%% file: sections/0_abstract.tex
\begin{abstract}
Learning visuomotor policies from scarce expert demonstrations remains a core challenge in robotic manipulation. A primary hurdle lies in distilling high-dimensional RGB representations into control-relevant geometry without overfitting.
While using frozen pretrained Vision Foundation Models (VFMs) improves data efficiency, it also shifts most task adaptation onto a small spatial pooling module, which can latch onto task-irrelevant shortcuts and lose geometric grounding when finetuned with few data samples.
More broadly, pretrained visual representations used for policy learning have been observed to struggle under even minor scene perturbations, highlighting the need for robustness-oriented inductive biases.
We propose \textbf{Geometric Anchor Pre-training (GAP)}, a simple, action-free warm-up stage that regularizes the spatial adapter \emph{before} downstream imitation learning.
GAP pre-trains the pooling layer on a lightweight simulated proxy task where object masks are available at no cost, encouraging the adapter to produce keypoints that lie on the object, cover its spatial extent (instead of collapsing), and remain sharp and repeatable over time.
This yields stable \emph{geometric anchors} that provide a reliable coordinate interface for few-shot policy learning, while keeping the VFM frozen.
We evaluate GAP on RoboMimic and ManiSkill under severe data scarcity (15--50 demonstrations) and domain shift.
A simple %
adapter regularized with GAP consistently outperforms stronger attention-based poolers and end-to-end fine-tuning, achieving 62\% success on RoboMimic \texttt{Can} with 15 demonstrations (+16\% over AFA), 63\% on the long-horizon high-precision \texttt{Tool Hang} task with 50 demonstrations ($+13\%$ over the best competitor based on R3M with Spatial Softmax) , and 61\% on ManiSkill \texttt{StackCube} with 30 demonstrations (+11\% over full fine-tuning).
The proxy stage is lightweight (about 40 minutes on a single consumer GPU) and fully decoupled from downstream tasks, making it practical to reuse across environments and manipulation skills. Project page: \hyperlink{https://lambdavi.github.io/gap/}{https://lambdavi.github.io/gap/}
\end{abstract}

%% file: sections/1_intro.tex
\section{Introduction}
Imitation Learning (IL) has become a strong paradigm for robotic manipulation, enabling robots to acquire complex, contact-rich behaviors directly from expert demonstrations.
Recent diffusion-based policies further improved performance by modeling multimodal action distributions and have become a common solution for visuomotor IL \cite{chi2023diffusion}.
However, in many practical settings, demonstrations are scarce, and learning reliable visuomotor policies remains challenging because the agent must extract task-relevant geometry from a few high-dimensional RGB observations without overfitting.

A common recipe to improve data efficiency is to (i) freeze a pretrained Vision Foundation Model (VFM) as the visual backbone (e.g., VC-1 \cite{majumdar2023vc1} or DINOv2 \cite{oquab2023dinov2}) and (ii) train a lightweight spatial bottleneck that compresses dense feature maps into a compact representation for control.
Methods using Diffusion Policies often instantiate this bottleneck with Spatial Softmax, yielding a set of 2D keypoints \cite{chi2023diffusion,finn2016deep}.
More semantic alternatives aim to learn which visual elements to retain: TokenLearner adaptively selects a small set of informative tokens \cite{ryoo2021tokenlearner}, while Attentive Feature Aggregation (AFA) is a lightweight trainable pooling mechanism designed to attend to task-relevant cues and suppress distractors without fine-tuning the frozen backbone \cite{tsagkas2025attentivefeatureaggregationor}.
These modular setups are appealing because they preserve the semantic knowledge of the pretrained VFM while keeping the task-specific learnable component small.

\begin{figure}[t]
    \centering
    \includegraphics[width=\columnwidth]{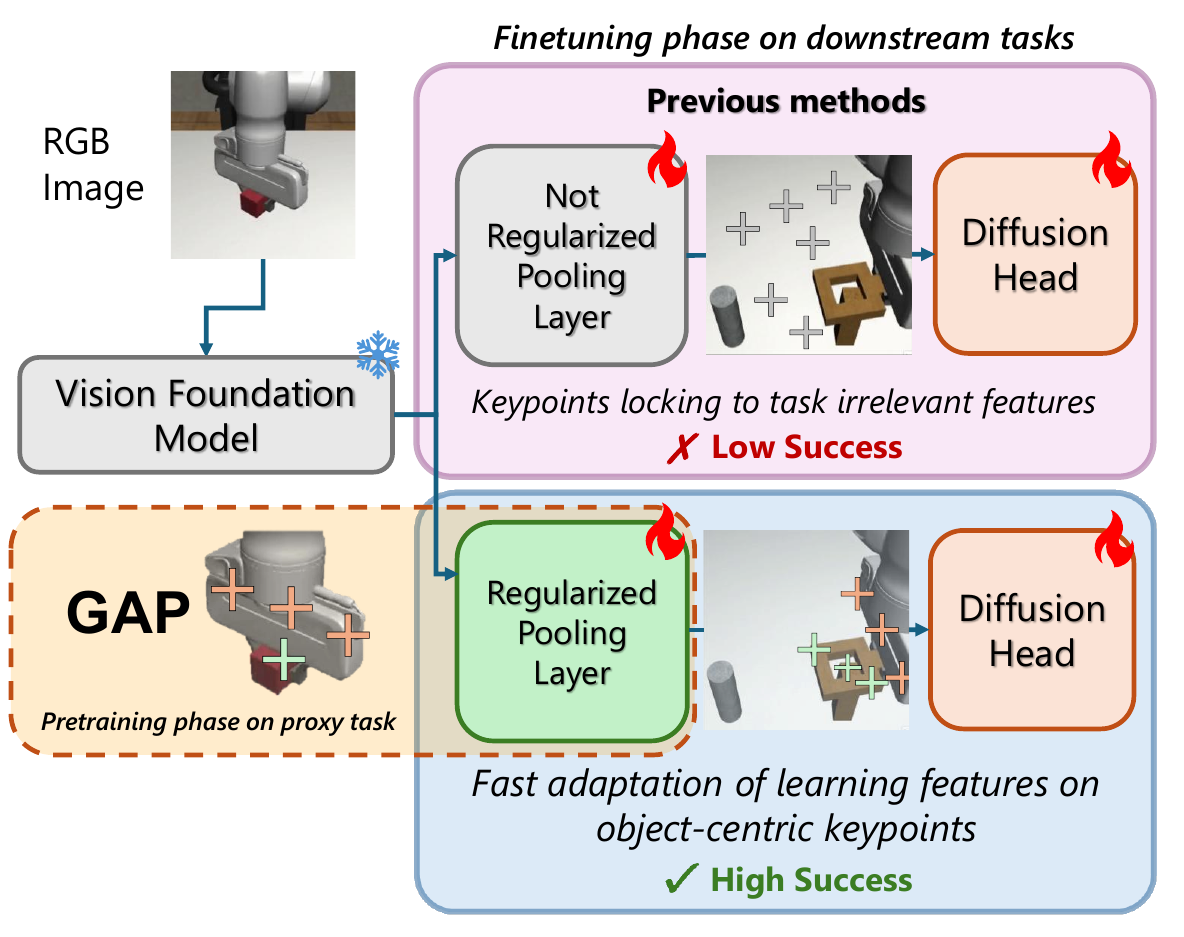}
    \caption{\textbf{Geometric Anchor Pretraining.} We introduce GAP, a pretraining strategy applied to the spatial pooling layer on a cheap proxy task. When only a few demonstrations are available for the target task ($N \le 50$), regularizing the spatial bottleneck with GAP consistently and largely outperforms other pooling techniques and end-to-end fine-tuning.}
    \label{fig:teaser}
\end{figure}

Yet, in the low-data regime, the spatial bottleneck is also the main point of failure.
With only a handful of demonstrations, the pooling module can lock onto easy-to-fit visual shortcuts (often in the background) instead of learning stable, object-centric geometry, producing keypoints that are poorly localized and brittle to minor test-time changes \cite{tsagkas2025temporaltrapentanglementpretrained}.
We refer to this failure mode as \emph{bottleneck collapse}: the adapter loses geometric grounding, and the downstream policy becomes unreliable under even small distribution shifts.

To address this, we propose \textbf{Geometric Anchor Pre-training (GAP)}, a simple yet particularly effective strategy that prevents bottleneck collapse by regularizing the spatial adapter \emph{before} downstream policy  (see Figure \ref{fig:teaser}).
The key intuition behind GAP is that many geometric cues needed for contact-rich manipulation---object extent, salient extremities, and stable spatial support---are transferable and can be learned without access to downstream actions.

GAP adds a short, action-free warm-up stage on a cheap simulated proxy task, where object masks are available at no cost.
During this warm-up, the adapter is encouraged to produce keypoints that lie on the object, cover it rather than collapsing to a single location, and remain sharp and repeatable across time.
This produces stable \emph{geometric anchors} that serve as a reliable coordinate interface when learning the downstream policy from a few demonstrations.

After this warm-up, we train the downstream policy in a few-shot setting (15--50 demonstrations), using the GAP-initialized adapter as the visual bottleneck.
We validate GAP on RoboMimic \cite{mandlekar2021matters} and ManiSkill \cite{tao2024maniskill3} under severe data scarcity and domain shift. Our results show satisfactory performance, with GAP achieving 62\% success on \texttt{Can} (RoboMimic) with only 15 demonstrations ($+16\%$ over AFA, $+7\%$ over full finetuning), 63\% on the high-precision long-horizon \texttt{Tool Hang} (RoboMimic) with 50 demonstrations ($+13\%$ over the best competitor R3M \cite{nair2023r3m} + Spatial Softmax \cite{finn2016deep}), and scoring 61\% on \texttt{StackCube} (Maniskill) with 30 demonstrations ($+11\%$ over full finetuning).
Notably, the proxy stage is lightweight (approximately 40 minutes on a single consumer GPU) and is fully decoupled from the downstream task, making it practical to reuse across tasks and environments.

To summarize, this paper contributes the following:
\begin{itemize}
    \item We introduce \textbf{GAP}, an action-free pretraining strategy for spatial pooling layers that injects geometric priors via a cheap mask-supervised proxy task, preventing bottleneck collapse in few-shot visuomotor IL.
    \item We demonstrate strong empirical gains in two different benchmarks (RoboMimic and ManiSkill) in extremely low-data regimes (15--50 demonstrations) under domain and tasks shift.%
    \item We provide experimental proofs that explicit geometric regularization is necessary, and that simply exposing the adapter to additional proxy data is insufficient and can cause negative transfer.
    \item We demonstrate that GAP learns transferable geometric anchors that improve robustness and data efficiency across tasks, visual domain shifts, and across simulation environments (pretrain on RoboMimic, transfer to ManiSkill).
\end{itemize}

%% file: sections/2_related.tex
\section{Related Work}
\label{sec:related_work}

\textbf{Visual Foundation Models for Robotics.} 
The adoption of large-scale pre-trained visual backbones has heavily driven recent advances in robot learning. Models such as MVP \cite{Xiao2022MVP}, VC-1 \cite{majumdar2023vc1}, and R3M \cite{nair2023r3m} have demonstrated that representations learned from internet-scale video datasets (e.g., Ego4D \cite{grauman2022ego4d}) or image-text pairs \cite{radford2021learning} can provide rich semantic features for downstream manipulation tasks. This paradigm is dominant because internet-scale pre-training provides robust generalization across novel object categories and diverse semantic environments.

However, this semantic focus introduces a fundamental trade-off: pre-training objectives that maximize semantic invariance (e.g., contrastive learning on image classification) inherently suppress high-frequency spatial and geometric details. In short, VFMs learn to recognize \emph{what} an object is, often at the expense of precisely locating \emph{where} its extremities are—a critical requirement for contact-rich manipulation. This comes with two major limitations. First, fine-grained tasks (e.g., \texttt{Tool Hang}, \texttt{Square Nut Assembly} \cite{mandlekar2021matters}) require sub-centimeter geometric precision that global semantic embeddings simply cannot resolve. Second, as demonstrated in our experiments, VFM-based policies frequently suffer visual distribution shifts \cite{zhuang2025enhancing} because they tend to rely on spurious correlations (e.g., background textures, lighting) rather than invariant geometric structures.

\label{sec:method}
\begin{figure*}[h!]
\centering
\includegraphics[width=0.92\linewidth]{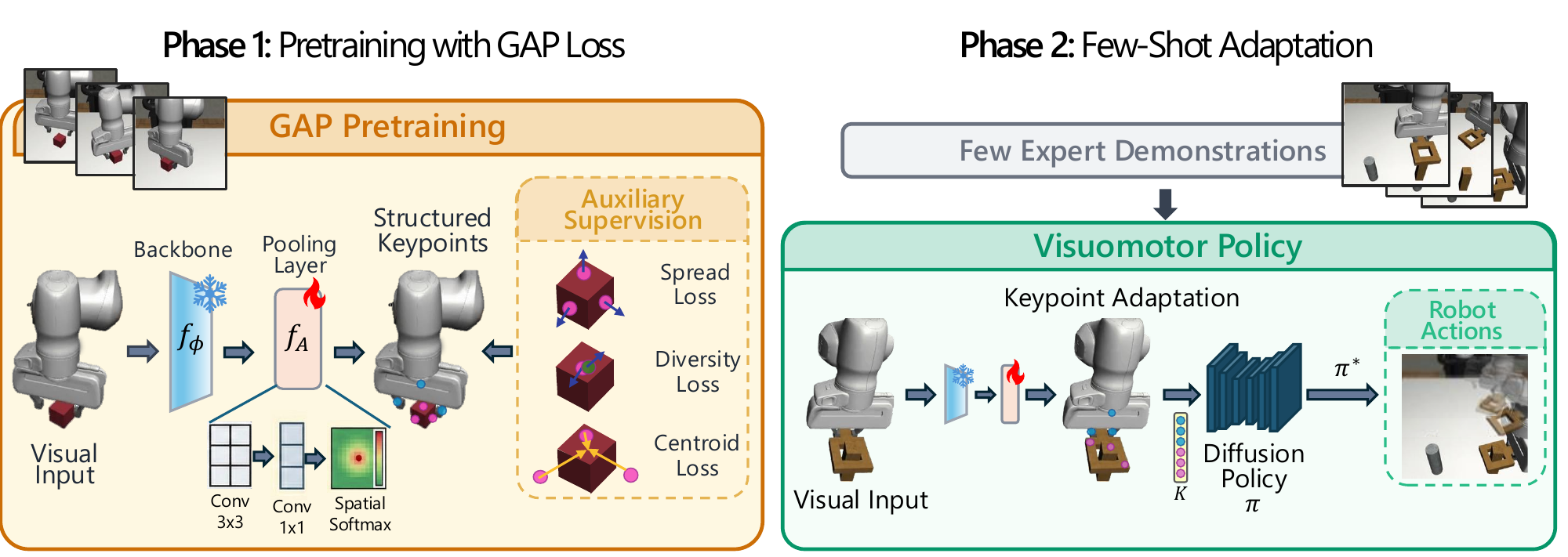}
\caption{\textbf{Method Overview.} 1. The spatial pooling layer extracts keypoints from the semantic pretrained backbone (frozen). GAP supervises this layer with the proposed loss, providing geometric grounding for policy learning. 2. Backbone and warmed-up pooling layer are then used to generate the input for the Diffusion Policy. During downstream training, the pooling layer is fine-tuned per task, to adapt object keypoints placement to the objects present in novel scenes.
}
\label{fig:method_overview}
\end{figure*}

\textbf{Keypoint-Based and Structured Representations.} 
Condensing high-dimensional visual states into sparse coordinate representations has a long history in robotics. Foundational architectures like Transporter Nets \cite{zeng2021transporter} and KeypointNet \cite{suwajanakorn2018discovery} extract low-dimensional structures via template matching or heuristic geometric reasoning. While sample-efficient, these early approaches often rely on reconstruction objectives that do not guarantee control-relevant disentanglement.

More recently, two divergent strategies have emerged. The first leverages large generative models—such as extracting features from diffusion U-Nets \cite{tang2023emergent} or utilizing VFMs for semantic correspondence \cite{wang2025skil}. While semantically rich, these methods introduce large computational overhead during inference or introduce dependencies on external reference images. The second strategy relies on structured geometric pipelines, pairing semantic keypoints with shape completion for category-level planning \cite{manuelli2019kpam, gao2021kpam}, or projecting depth into structured point maps \cite{jia2025pointmappolicy}. 

These diverse approaches confirm a broader consensus: visuomotor control fundamentally benefits from explicit spatial structure. However, existing methods force a compromise, requiring either heavy inference-time computation, external model dependencies, or dense manual supervision. GAP avoids this compromise. By enforcing geometric consistency through a synthetic, mask-supervised proxy task, GAP makes spatial anchor extraction lightweight, reference-free, and seamlessly scalable to standard end-to-end Diffusion Policies.

\textbf{Pooling strategies for Vision Foundation Models}
As Vision Foundation Models (VFMs) output dense, high-dimensional feature maps, compressing this spatial information for downstream policy learning has become a critical architectural focus. Recent approaches leverage attention mechanisms to dynamically aggregate visual information into compact token representations. Methods such as TokenLearner \cite{ryoo2021tokenlearner} and Perceiver IO \cite{jaegleperceiver} reduce spatial dimensionality by computing attention between learned latent queries and the input feature map, theoretically allowing the network to focus only on task-relevant semantic regions. Building on these, in the context of robotic manipulation, frameworks employing Attention Feature Aggregation (AFA) \cite{tsagkas2025attentivefeatureaggregationor} and similar transformer-based bottlenecks have been proposed to filter visual distractors and improve robustness by explicitly learning only semantic features that are important for the manipulation tasks.
However, these semantic attention mechanisms rely heavily on larger demonstration corpora to learn robust query-key mappings. As we demonstrate empirically, this high capacity becomes a critical liability in severe low-data regimes (e.g., 15--50 demonstrations), leading to severe representation collapse. Rather than isolating precise object coordinates, highly parameterized attention poolers frequently overfit to spurious transient features, such as background textures or specific lighting conditions. GAP explicitly counters this by trading semantic flexibility for strict geometric equivariance. By forcing the VFM features through a low-parameter, rigidly regularized spatial adapter, GAP extracts stable $(x, y)$ keypoints instead of diffuse attention maps. This strict structural constraint prevents the texture-hijacking typical of high-capacity poolers, providing robust and noise-free spatial priors that largely accelerate convergence in data-starved environments while providing a simpler representation to the policy .

%% file: sections/3_method.tex
\section{Methodology}
\textbf{Geometric Anchor Pretraining (GAP)} is a pretraining strategy designed to extract object-centric geometric priors for visuomotor imitation learning directly from dense VFM's embeddings. GAP addresses the spatial overfitting commonly observed in data-scarce regimes ($N \le 50$) by explicitly supervising a coordinate-based adapter via a masked proxy task. %
We begin by formalizing the imitation learning setting and defining the spatial adapter architecture. We then introduce our method (Figure \ref{fig:method_overview}), focusing on the proxy task and losses, followed by a description of the downstream policy adaptation procedure.

\subsection{Imitation Learning with Diffusion Policies}
\label{subsec:background}
We focus on visuomotor policy learning in the Imitation Learning setting, where an agent learns control behaviors from expert demonstrations. The agent is provided with a dataset $\mathcal{D} = \{ \tau_i \}_{i=1}^N$, where each trajectory $\tau_i = \{ (o_t, a_t) \}_{t=0}^T$ consists of visual observations $o_t$ and corresponding actions $a_t$. The objective is to infer a policy $\pi_\theta(a_t | o_t)$ that reproduces the expert behavior.

In this work, we adopt \textbf{Diffusion Policies} \cite{chi2023diffusion}, which parameterize the action distribution $\pi_\theta(a_t | o_t)$ as a conditional denoising diffusion probabilistic model. The training objective minimizes the noise prediction error:
\begin{equation}
    \mathcal{L}_{diff} = \mathbb{E}_{\epsilon \sim \mathcal{N}, t \sim \mathcal{U}, \tau \sim \mathcal{D}} \left[ \| \epsilon - \epsilon_\theta(a_t^{(k)}, k, E(o_t)) \|_2^2 \right]
\end{equation}
where $E(o_t)$ is the visual embedding used to condition the denoising network $\epsilon_\theta$ at diffusion step $k$. While a wide range of Imitation Learning algorithms have been introduced, with no loss of generality, in this work, we experiment with \cite{chi2023diffusion} as the policy learning method for all the experiments to isolate the impact of the conditioning representation. GAP focuses entirely on the pre-training of the adapter between the frozen vision encoder and the policy $\pi_\theta$ to provide a robust, geometry-aware conditioning signal.

\subsection{The Spatial Adapter}
\label{subsec:arch}
To extract robust semantic features, we use a frozen pretrained backbone $f_\phi$ (e.g., ResNet-50, ViT-S, or ViT-B). To map visual features into precise spatial embeddings, $f_\phi$ is followed by a lightweight adapter module, which we denote as $f_A$ (see \autoref{fig:method_overview}). Specifically, $f_A$ first applies a shallow convolutional network (a $3\times3$ followed by a $1\times1$ convolution) to project the high-dimensional backbone features into $K$ spatial activation maps, denoted as $\Phi_t \in \mathbb{R}^{K \times h \times w}$. Finally, $f_A$ applies a Spatial Softmax (SS) \cite{finn2016deep} module to convert these maps into $K$ 2D spatial coordinates, which we define as our candidate \emph{keypoints} $P_t = \{p_{k,t}\}_{k=1}^K$:

\begin{equation}
    p_{k,t} = \sum_{x=1}^{w} \sum_{y=1}^{h} \begin{bmatrix} x \\ y \end{bmatrix} \frac{\exp(\Phi_{t, k, x, y})}{\sum_{x',y'} \exp(\Phi_{t, k, x', y'})}
\end{equation}
where $\Phi_{t, k, x, y}$ is the activation of the $k$-th feature channel at spatial location $(x,y)$. 

This mapping converts the visual learning paradigm into a state-based one, compressing dense feature maps into $K$ sparse 2D coordinates. When training policies for 200+ demonstrations, we observe that the learned keypoints naturally stick to specific objects and become reliable ``semantic trackers''. However, under extremely low-data regime, without explicit supervision, these keypoints tend to latch onto spurious visual cues rather than geometrically meaningful locations. %
This occurs because the downstream action regression loss ($\mathcal{L}_{diff}$) provides only weak, indirect spatial supervision. Forced to minimize training error with minimal data, the network takes a shortcut, anchoring to high-contrast static distractors (e.g., table textures) instead of complex object geometry.

\subsection{Geometric Anchor Pretraining (GAP)}
\label{subsec:gap_method}
To address this, GAP pretrains the spatial bottleneck $f_A$ on a single, cheap simulated proxy task. This aims to decouple geometric feature learning from action-mapping.

\begin{figure}[ht!]
\centering
\includegraphics[width=0.85\columnwidth]{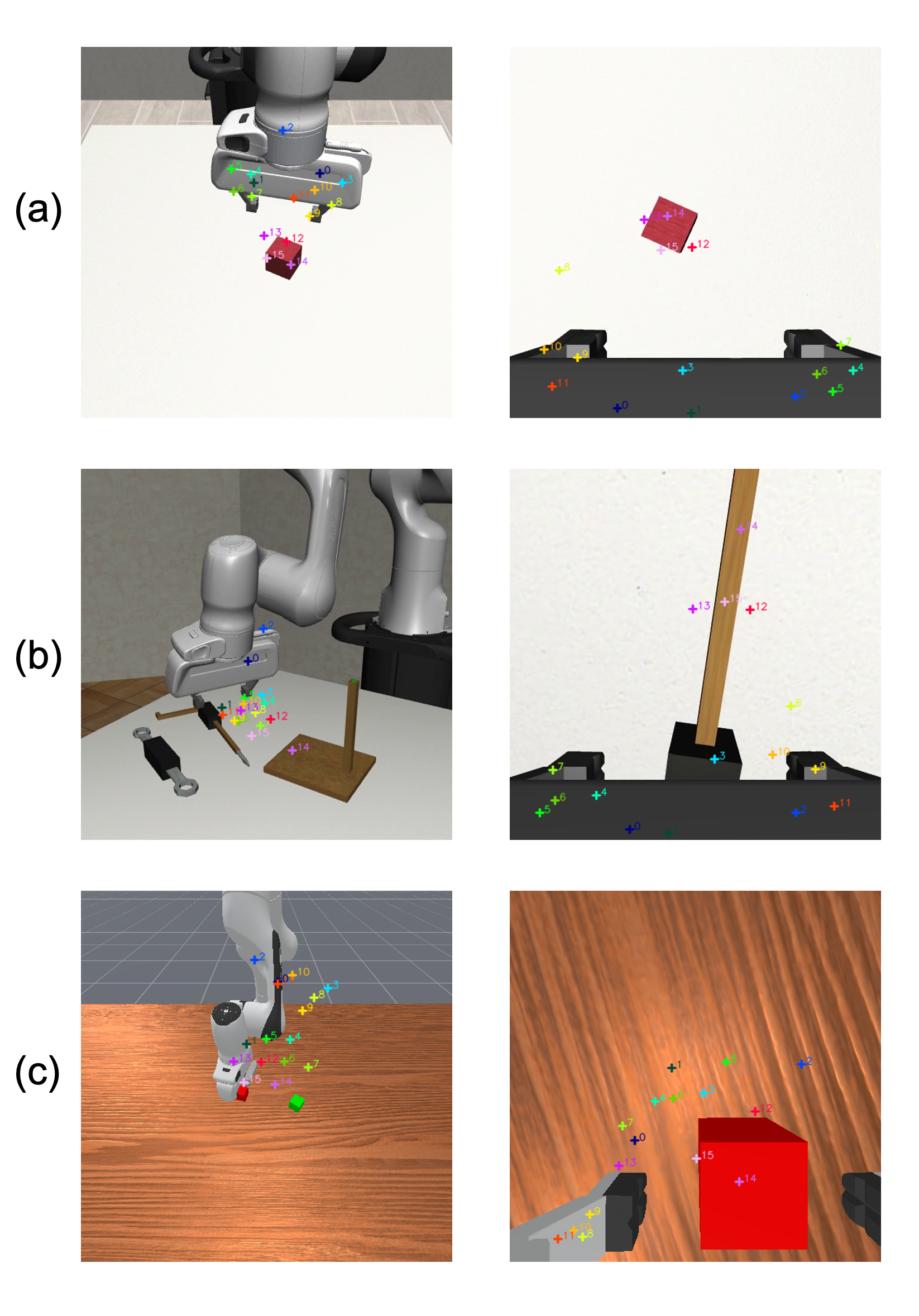}
\caption{\textbf{Qualitative Keypoint Transfer.} When pre-trained on a proxy task and then transferred to a new task and simulator, GAP allows for keeping a favorable geometric grounding even in zero-shot. \textbf{(a)} shows the keypoints placement on the Robomimic \texttt{LiftCube} task after pretraining. \textbf{(b)} and \textbf{(c)} show keypoints positioning when using the pre-trained visual encoder in zero-shot on a different task (of the same simulator) and on a different task and simulator (Maniskill), respectively. We show a third-person view on the left and a first-person view on the right.}
\label{fig:qualitative_small}
\vspace{-3pt}
\end{figure}

\textbf{Proxy Task.} 
The objective of our training is to align geometric keypoints with task-relevant objects in the scene, without any task-specific knowledge. Therefore, in principle we can use for pretraining any simple manipulation task or contact-rich motor babbling. In this paper, with no loss of generality, we experiment with the \texttt{LiftCube} task from Robomimic \cite{mandlekar2021matters}---
the simplest task available in the benchmark---in which a Franka Emika Panda robot is tasked to reach, grasp, and lift a randomly positioned cube on a plain \emph{white-background} table (see \autoref{fig:qualitative_small}-a). To evaluate proxy task invariance, we also experiment with a different pre-training task (PlaceSphere from ManiSkill).
Trajectories are generated automatically via a scripted controller, requiring no human teleoperation. We use 100 demonstrations of this proxy task, leveraging ground-truth object segmentation masks $\mathcal{M}_t$ provided by the simulator to supervise our loss. Crucially, no expert action labels $a_t$ are needed at any point during this phase. %

\textbf{GAP Spatial Objectives.}
We supervise the adapter $f_A$ using a multi-objective spatial loss that enforces object-centric, spatially distributed, and non-redundant keypoints over the robot gripper and interacting objects. This is achieved through three components, depicted in \autoref{fig:method_overview} and detailed below.

\subsubsection{Centroid Alignment ($\mathcal{L}_{center}$)}
To ensure keypoints ground themselves on the target object rather than background distractors, we minimize the distance between the predicted keypoint centroid $\bar{p}_t$ and the ground-truth mask centroid $c_t$:
\begin{equation}
    \mathcal{L}_{center} = \| \bar{p}_t - c_t \|_2^2 \quad \text{where} \quad \bar{p}_t = \frac{1}{K} \sum_{k=1}^K p_{k,t}
\end{equation}
and $c_t$ is computed via the spatial moments of the binary mask $\mathcal{M}_t$.

\subsubsection{Geometric Spread ($\mathcal{L}_{spread}$)}
To prevent the degenerate solution where all keypoints collapse precisely in the centroid (which drops orientation information), we enforce the spatial variance of the keypoints $\sigma_p$ to match the normalized object scale $\sigma_{target}$:
\begin{equation}
    \mathcal{L}_{spread} = \| \sigma_p - \sigma_{target} \|_2^2 \quad \text{where} \quad \sigma_p = \frac{1}{K} \sum_{k=1}^K \| p_{k,t} - \bar{p}_t \|_2
\end{equation}
The target scale is derived from the mask area $A_t = \sum \mathcal{M}_t$, approximated as $\sigma_{target} = 0.8 \times \sqrt{A_t / \pi}$, representing a proportional bounding radius.

\subsubsection{Keypoint Diversity ($\mathcal{L}_{div}$)}
Lastly, to maximize the structural information captured by the bottleneck, we penalize redundancy by enforcing a minimum separation margin $\delta_{min}$ between any pair of keypoints:
\begin{equation}
    \mathcal{L}_{div} = \frac{1}{K} \sum_{k=1}^K \left[ \max\left(0, \delta_{min} - \min_{j \neq k} \| p_{k,t} - p_{j,t} \|_2 \right) \right]^2
\end{equation}
This term encourages the network to discover the object's distinct geometric extremities, thereby generating highly informative \emph{Geometric Anchors}.

The final pre-training objective combines these terms to create a "push-pull" dynamic: keypoints are pulled onto the object ($\mathcal{L}_{center}$), but pushed outward to span its geometry ($\mathcal{L}_{spread}$) and away from one another ($\mathcal{L}_{div}$).
\begin{equation}
    \mathcal{L}_{GAP} = \lambda_c \mathcal{L}_{center} + \lambda_s \mathcal{L}_{spread} + \lambda_d \mathcal{L}_{div}
\end{equation}
In our implementation, we prioritize spatial coverage over strict centering by setting $\lambda_c = 0.3$, $\lambda_s = 0.5$, and $\lambda_d = 2.0$, with a diversity margin $\delta_{min} = 0.15$ (normalized image coordinates). We report an extensive ablation on the different objective components in Sec. \ref{ablation_losses}.

\indent\textbf{Object-Centric Keypoint Allocation.} While end-to-end policies require massive datasets to naturally develop entity-centric keypoints, GAP explicitly enforces this optimal behavior in low-data regimes. Given a pre-training scene with $M$ semantic entities, we partition the $K$ available keypoints into $M$ disjoint subsets: $P_t = \bigcup_{m=1}^{M} P_{t,m}$. The spatial regularization objectives ($\mathcal{L}_{GAP}$) are then applied independently to each subset using its corresponding mask. 
This $M$-way partition bridges the abstract semantic expressivity of Vision Foundation Models (VFMs) with the strict, object-centric geometric priors required for physical manipulation. Consequently, downstream fine-tuning does not need to learn object separation from scratch. When transitioning from a proxy task to novel, multi-object environments (e.g., \texttt{StackCube}, \texttt{SquareNut}), these pre-trained subsets deploy as independent semantic trackers. Having already internalized priors of centroid alignment, spatial spread, and extremity-seeking diversity, they rapidly re-anchor to novel geometries with minimal adaptation. \autoref{fig:qualitative_small} qualitatively demonstrates this robust, entity-separated initialization prior to any policy training.

\subsection{Downstream Policy Adaptation}
\label{subsec:downstream_adapt}
Following GAP, the regularized adapter and frozen encoder $E$ are transferred to the downstream tasks. The $M$ pre-trained keypoint subsets drastically reduce the policy's learning burden: subsets tracking persistent elements (e.g., the manipulator) require light adaptation, allowing the few-shot demonstrations to focus entirely on grounding the remaining keypoints to novel target objects. This preserves learned spatial priors and enables highly sample-efficient convergence.
For fair comparison, \textbf{all evaluated baselines}---including end-to-end models and VFMs (R3M, DINOv2, VC-1)---employ an identical architecture. Models denoted with ``SS'' utilize our convolutional pooler and Spatial Softmax. During downstream fine-tuning, privileged segmentation masks $\mathcal{M}_t$ are strictly discarded. The vision backbone remains frozen, and we fine-tune only the lightweight adapter $f_A$ alongside the diffusion head via the action-prediction objective $\mathcal{L}_{diff}$.

%% file: sections/4_exps.tex
\section{Experiments}
\label{sec:exp}

We evaluate GAP on two simulation benchmarks:  Robomimic \cite{mandlekar2021matters} and ManiSkill3 \cite{tao2024maniskill3}. For all tasks, our observations consist of two camera views (i.e., an agent-centric and a wrist-mounted camera). Each camera stream is processed independently through its own instantiated visual backbone and GAP-regularized adapter $f_A$. The resulting keypoints are then concatenated before being passed to the diffusion policy. 
Noteworthy, we measure the number of trainable parameters of the methods: End-to-end training of the Resnet50 requires updating about 56M of parameters, while AFA and GAP require about 3M and 2M, respectively.
Our experimental evaluation is designed to answer three research questions:
1) Does explicit geometric pre-training of the spatial bottleneck (GAP) improve data efficiency in the $N \leq 50$ regime, and does the learned prior transfer across simulator and tasks?
2) Is GAP appropriate to mitigate the ''bottleneck`` problem across backbones architecture and VFMs?
3) What drives GAP's performance gain: the proxy data, the geometric loss, or the disentangled keypoint structure?
\vspace{-2pt}
\subsection{Experimental Setup}

\input{sections/main_results_table}

We evaluate on four tasks of increasing difficulty: \texttt{PickAndPlace Can}, \texttt{Square Nut Assembly}, and \texttt{Tool Hang} from Robomimic \cite{mandlekar2021matters}, and \texttt{StackCube} from ManiSkill3 \cite{tao2024maniskill3}. \texttt{PickAndPlace Can} is a relatively simple pick-and-place task; \texttt{Square Nut Assembly} requires precise peg insertion; \texttt{Tool Hang} is a long-horizon, multi-step assembly task; \texttt{StackCube} requires stacking one cube precisely above another. The difficulty of the task comes from the heavy randomization of object positioning over the whole table, which represents a non-trivial problem in a data-scarce regime. All tasks are evaluated over different settings, using 15, 20, 30, and 50 expert demonstrations available for policy training. Results are averaged over three seeds.

\textbf{Baselines.} For each task and setting we compare: a Resnet50 End-to-End full fine-tuned (E-E) replicating a Diffusion Policy setup \cite{chi2023diffusion}, three frozen VFM backbones paired with a standard Spatial Softmax-based adapter (R3M+SS, DINOv2+SS, VC-1+SS), and the best performing backbone paired with Attention Feature Aggregation (AFA) \cite{tsagkas2025attentivefeatureaggregationor} as SOTA attention pooler. DinoV2 is used with their ViT-S backbone, and VC-1 is used with their ViT-B to study different backbone sizes/pretraining impact on downstream policy learning.
All baselines share the same downstream architecture (U-Net trained with diffusion objective); leaving as only difference between them the design of the pooling layer and how it is trained.
We train all models for 1,000 epochs with 512 as batch size, and report the results with their best performing learning rates (between 1e-3 and 1e-5) and their best configurations (backbone) in the case of AFA and GAP. For all experiments we fix the number of keypoints to 16 per camera.
For all tasks and settings, we use the same pre-training proxy task, i.e., \texttt{LiftCube} from Robomimic, scripted automatically on a white-background table, requiring no expert action labels (Figure \ref{fig:qualitative_small}). Pre-training runs for $10{,}000$ steps ($\approx$40 min on one A40 GPU) and relies solely on simulator-provided segmentation masks, which are 
used only in this pre-training phase.

\subsection{Impact of Pretraining on Downstream Task Learning}
\input{sections/4_3table}
Table~\ref{table:main_results} reports success rates across all four tasks and settings. We note that semantic attention (AFA) \cite{tsagkas2025attentivefeatureaggregationor} overfits heavily at low demo counts: e.g. on StackCube using 15 demos, AFA achieves only $0.09$ while VC-1+SS reaches $0.04$ and GAP reaches $\mathbf{0.20}$. With 30 demos, AFA ($0.25$) is actually outperformed by a simpler VC-1+SS ($0.28$), confirming that larger pooling capacity may be harmful when data is scarce; GAP performs two times better achieving $\mathbf{0.61}$. We also observe that unregularized SS yields inconsistent results, performing reasonably well on simple tasks but failing to anchor to object geometry on harder ones. In addition, it is worth noting that E-E fine-tuning underperforms on \texttt{Tool Hang} and lags behind GAP on every task, despite updating more parameters.
GAP consistently achieves the highest success rate across all tasks and settings. On \texttt{Can}, GAP reaches $0.62$ with only 15 demos versus AFA's $0.46$ and E-E's $0.55$, and reaches $0.96$ at 50 demos. On the challenging \texttt{Square} task, GAP scores $0.53$ with 50 demos versus $0.43$ for AFA and $0.38$ for E-E. On \texttt{Tool Hang} E-E fails in learning the task with 15 demos, while GAP achieves a reasonable $0.27$, and $0.63$ using 50 demos, consistently outperforming all baselines.
Finally, we evaluate the \texttt{Square} task using 100 demonstrations to observe if abundant expert data mitigates spatial bottleneck collapse. At this scale, VC-1 + SS, VC-1 + AFA, and our GAP framework achieve success rates of $0.64 \pm 0.05$, $0.65 \pm 0.03$, and $0.68 \pm 0.02$, respectively. As expected, the massive influx of expert trajectories allows the unregularized baselines to largely close the performance difference, confirming that GAP's geometric prior is most critical—and provides the highest relative gains—in data-scarce regimes, but still provides advantages in higher ones.
A natural concern is that GAP's advantage may stem from extra data in the training domain rather than a genuinely general geometric prior. Table~\ref{table:domain_ablation} reports an ablation study to verify it. We compare three pre-training conditions on StackCube: \textit{Same Table} ({PlaceSphere} in ManiSkill as the proxy task, using the same wooden table as the downstream task), \textit{Cross Table} (modified {PlaceSphere} with white
table background on Maniskill as proxy tasks, visual shift), and \textit{Cross-Sim} (\emph{LiftCube} in Robomimic as proxy task, simulator shift). At 30 demos, the three conditions yield on average $0.56$, $0.54$, and $0.59$, respectively---with no statistically significant difference between them. All three variations largely outperform the VC-1+SS baseline ($0.28$). This confirms that GAP learns general priors rather than simulator-specific textures or visual priors.
We also note that the success of \textit{Cross-Sim} also proves that GAP is task-agnostic to the choice of proxy task.

\subsection{GAP impact on different backbones}
\input{sections/figure_abl_bakbones}
To prove that bottleneck collapse is a problem for various backbones, Figure \ref{fig:backbone_abl} evaluates all pooling strategies across R3M, frozen DINOv2, and frozen VC-1 on the Square task at 30 demos, requiring learning of precise actions from few demos. Additionally, we also present the results on StackCube (Maniskill). Similar trends can be observed for other tasks and are here omitted for the sake of space. The results demonstrate a flaw in semantic pooling: adding AFA to DINOv2 \emph{degrades} performance w.r.t. standard SS ($0.19$ vs.\ $0.23$), confirming that semantic mechanisms are not enough when data is scarce. GAP instead improves DINOv2, bringing it from $0.23$ to $\mathbf{0.29}$, and pushes VC-1 to the state-of-the-art $\mathbf{0.37}$ on this task---higher than any other method including E-E fine-tuning ($0.29$). GAP acts as a spatial regularizer, compatible with any backbone.

\input{sections/loss_ablation_table}

\input{sections/pretr_ablation}
\subsection{Ablation on GAP objective} \label{ablation_losses}
We ablate the three components of the GAP loss on the Square task with 30 demos. Numerical results are reported in Table~\ref{table:loss_ablation}. Removing $\mathcal{L}_{div}$ causes all keypoints to collapse into the centroid of masks, which causes the loss of orientation and boundary information. Removing $\mathcal{L}_{spread}$ prevents keypoints from reaching the object's geometric boundaries, resulting in a tightly clustered, brittle representation. Both degradations reduce downstream policy success, confirming that all three loss terms---centroid alignment, spread, and diversity---are strictly necessary to fully regularize the bottleneck.

Finally, we also test whether GAP's effectiveness stems from the geometric structure of the proposed loss, or merely from exposure to additional proxy data. While the cross-simulator experiment already rules out simulator-specific data leakage as the primary driver of performance, it does not control for the total amount of proxy data seen. To isolate the contribution of the pre-training objective itself, we compare GAP against baselines that are pre-trained on the same proxy demonstrations but in an end-to-end fashion with a diffusion head.

Specifically, we pre-train both the unregularized bottleneck (Spatial Softmax) and AFA directly on the proxy task using action-supervised imitation, giving them access to the same $100$ proxy demonstrations used by GAP — plus the expert action labels and full simulator state, which GAP never requires. This setup deliberately provides a favorable advantage to the baselines, ensuring that any performance gap can be attributed to the geometric pre-training objective rather than data quantity or domain exposure.

Importantly, in this experiment, we only transfer the pre-trained 
vision encoder $f_\phi$ and the adapter $f_A$, deliberately withholding the pretrained diffusion policy, in order to isolate the contribution of the 
pooling layer pre-training.

Table \ref{table:pretrain_abl} shows an interesting result about the failure mode of standard pretraining. Although all the baselines reach 100\% success rate in the trivial \textit{Lift} task, when transferred to the downstream task, the pooling layer fails to adapt. GAP, in contrast, fully exploits its pretraining, obtaining a 12\% improvement over the baseline.

\begin{figure}[h]
    \centering
    \includegraphics[width=\linewidth]{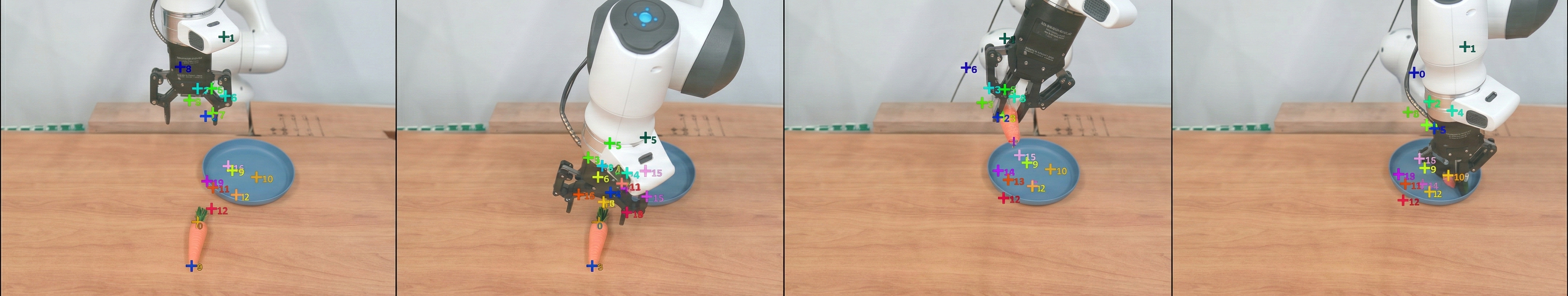}
    \caption{\textbf{VC-1 backbone with GAP pretrained spatial pooler in the wild.} We use one video from \cite{fang2025rebot} of a real-world robot performing a pick and place task and apply the GAP pretrained model (Lift task of Robomimic) to the video. This qualitatively shows a very good initialization of the keypoints even in the sim-to-real scenario, which vouches for good transfer of results in the real world.}
    \label{fig:real_world}
\end{figure}

While real-world tasks are not analyzed in this work for the sake of space, and are left for extension, we present a qualitative snapshot depicting zero-shot application of GAP pretraining to a real-world video (no policy learning) in Figure \ref{fig:real_world}.

%% file: sections/main_results_table.tex
\begin{table*}[t]
\centering
\sffamily 
\renewcommand{\arraystretch}{1.1}

\resizebox{\textwidth}{!}{
\begin{tabular}{l ccccc ccccc}
\toprule
& \multicolumn{5}{c}{\textbf{Robomimic: Can}} & \multicolumn{5}{c}{\textbf{Robomimic: Square}} \\
\cmidrule(lr){2-6} \cmidrule(lr){7-11}
\textbf{Method} & \textbf{15} & \textbf{20} & \textbf{30} & \textbf{50} & \textbf{Avg} & \textbf{15} & \textbf{20} & \textbf{30} & \textbf{50} & \textbf{Avg} \\
\midrule
E-E (Full FT) & \res{0.55}{0.04} & \res{0.76}{0.02} & \res{0.88}{0.03} & \res{0.95}{0.01} & \avg{0.785} & \res{0.15}{0.03} & \res{0.19}{0.02} & \res{0.29}{0.03} & \res{0.38}{0.02} & \avg{0.253} \\
R3M + SS & \res{0.50}{0.02} & \res{0.75}{0.05} & \res{0.78}{0.04} & \res{0.86}{0.02} & \avg{0.723} & \res{0.12}{0.02} & \res{0.13}{0.04} & \res{0.17}{0.04} & \res{0.22}{0.02} & \avg{0.158} \\
DINOv2 + SS & \res{0.51}{0.08} & \res{0.68}{0.03} & \res{0.73}{0.03} & \res{0.86}{0.00} & \avg{0.695} & \res{0.10}{0.00} & \res{0.22}{0.02} & \res{0.23}{0.04} & \res{0.26}{0.03} & \avg{0.203} \\
VC-1 + SS & \res{0.49}{0.06} & \res{0.64}{0.10} & \res{0.82}{0.05} & \res{0.89}{0.02} & \avg{0.710} & \res{0.07}{0.06} & \res{0.24}{0.06} & \res{0.23}{0.03} & \res{0.32}{0.05} & \avg{0.215} \\
AFA  & \res{0.46}{0.01} & \res{0.74}{0.02} & \res{0.78}{0.02} & \res{0.93}{0.05} & \avg{0.728} & \res{0.15}{0.02} & \res{0.19}{0.03} & \res{0.32}{0.04} & \res{0.43}{0.04} & \avg{0.273} \\
\rowcolor{TechTeal} \textbf{GAP (Ours)} & \best{0.62}{0.06} & \best{0.80}{0.02} & \best{0.94}{0.04} & \best{0.96}{0.02} & \bestavg{0.830} & \best{0.20}{0.03} & \best{0.33}{0.03} & \best{0.37}{0.01} & \best{0.53}{0.03} & \bestavg{0.358} \\
\addlinespace[1ex] 
& \multicolumn{5}{c}{\textbf{Robomimic: Tool Hang}} & \multicolumn{5}{c}{\cellcolor{SimGray}\textbf{ManiSkill: StackCube}} \\
\cmidrule(lr){2-6} \cmidrule(lr){7-11}
\textbf{Method} & \textbf{15} & \textbf{20} & \textbf{30} & \textbf{50} & \textbf{Avg} & \cg\textbf{15} & \cg\textbf{20} & \cg\textbf{30} & \cg\textbf{50} & \cg\textbf{Avg} \\
\midrule
E-E (Full FT) & \res{0.06}{0.05} & \res{0.2}{0.1} & \res{0.13}{0.06} & \res{0.33}{0.06} & \avg{0.18} & \cg\best{0.22}{0.10} & \cg\res{0.23}{0.05} & \cg\res{0.50}{0.08} & \cg\res{0.66}{0.08} & \cg\avg{0.403} \\
R3M + SS & \res{0.16}{0.06} & \res{0.17}{0.06} & \res{0.27}{0.06} & \res{0.50}{0.10} & \avg{0.275} & \cg\res{0.06}{0.01} & \cg\res{0.09}{0.02} & \cg\res{0.15}{0.05} & \cg\res{0.38}{0.09} & \cg\avg{0.171} \\
DINOv2 + SS & \res{0.23}{0.06} & \res{0.13}{0.06} & \res{0.20}{0.17} & \res{0.23}{0.23} & \avg{0.198} & \cg\res{0.07}{0.01} & \cg\res{0.10}{0.03} & \cg\res{0.15}{0.00} & \cg\res{0.60}{0.05} & \cg\avg{0.230} \\
VC-1 + SS & \res{0.20}{0.05} & \res{0.17}{0.11} & \res{0.20}{0.10} & \res{0.43}{0.05} & \avg{0.250} & \cg\res{0.04}{0.01} & \cg\res{0.08}{0.05} & \cg\res{0.28}{0.02} & \cg\res{0.63}{0.03} & \cg\avg{0.258} \\
AFA  & \res{0.20}{0.10} & \res{0.23}{0.10} & \res{0.30}{0.06} & \res{0.45}{0.10} & \avg{0.295} & \cg\res{0.09}{0.02} & \cg\res{0.14}{0.03} & \cg\res{0.25}{0.02} & \cg\res{0.44}{0.08} & \cg\avg{0.230} \\
\rowcolor{TechTeal} \textbf{GAP (Ours)} & \best{0.27}{0.06} & \best{0.33}{0.06} & \best{0.37}{0.06} & \best{0.63}{0.05} & \bestavg{0.400} & \res{0.20}{0.03} & \best{0.24}{0.04} & \best{0.61}{0.10} & \best{0.80}{0.02} & \bestavg{0.463} \\
\bottomrule
\end{tabular}
}
\caption{\textbf{Multi-Task Evaluation Results.} For all tasks, we pre-train on \textit{LiftCube} from Robomimic. Results on the ManiSkill simulator environment are shaded in gray to denote the domain shift. GAP achieves state-of-the-art average performance. For GAP the best performing VFM is VC1 with ViT-B while for AFA is VC1 for \textit{Can} and R3M for the other tasks.}
\label{table:main_results}
\end{table*}

%% file: sections/4_3table.tex
\begin{table}[t]
\centering
\sffamily 
\renewcommand{\arraystretch}{1.2} 
\resizebox{0.95\columnwidth}{!}{
\begin{tabular}{@{}lcccc@{}}
\toprule
\textbf{Pre-training Strategy} & \textbf{15 Demos} & \textbf{20 Demos} & \textbf{30 Demos} & \textbf{50 Demos} \\ 
\midrule

Resnet50 E-E (\cite{chi2023diffusion}) \ & \res{0.22}{0.10} & \res{0.23}{0.05} & \res{0.50}{0.08} & \res{0.66}{0.08} \\
VC-1 + SS (No Pre-training) & \res{0.04}{0.01} & \res{0.08}{0.05} & \res{0.28}{0.02} & \res{0.63}{0.03} \\ 
AFA & \res{0.09}{0.02} & \res{0.14}{0.03} & \res{0.25}{0.02} & \res{0.44}{0.08} \\ 
\midrule
\rowcolor{TechTeal} \textbf{GAP (Same Table)} & \best{0.36}{0.07} & \best{0.35}{0.06} & \res{0.56}{0.10} & \best{0.80}{0.01} \\
\rowcolor{TechTeal} \textbf{GAP (Cross Table)} & \res{0.28}{0.15} & \res{0.31}{0.06} & \res{0.54}{0.10} & \res{0.74}{0.02} \\
\rowcolor{TechTeal} \textbf{GAP (Cross-Sim)} & \res{0.19}{0.06} & \res{0.24}{0.04} & \best{0.61}{0.1} & \best{0.80}{0.02} \\ 

\bottomrule
\end{tabular}
}
\caption{\textbf{Proxy Task Ablation. (StackCube, Maniskill)} The performance of StackCube on Maniskill are analyzed under three pretrained conditions: Same Table (pretraining on \textit{PlaceSphere} on Maniskill), Cross Table (pretraining on modified \textit{PlaceSphere} with white table background on Maniskill), and Cross-Sim (pretraining on \textit{LiftCube} on Robomimic). Performance remains consistent across all settings, demonstrating that GAP learns domain-agnostic geometric priors.}
\label{table:domain_ablation}
\end{table}

%% file: sections/figure_abl_bakbones.tex
\begin{figure*}[h!]
    \centering
    \includegraphics[width=0.8\linewidth]{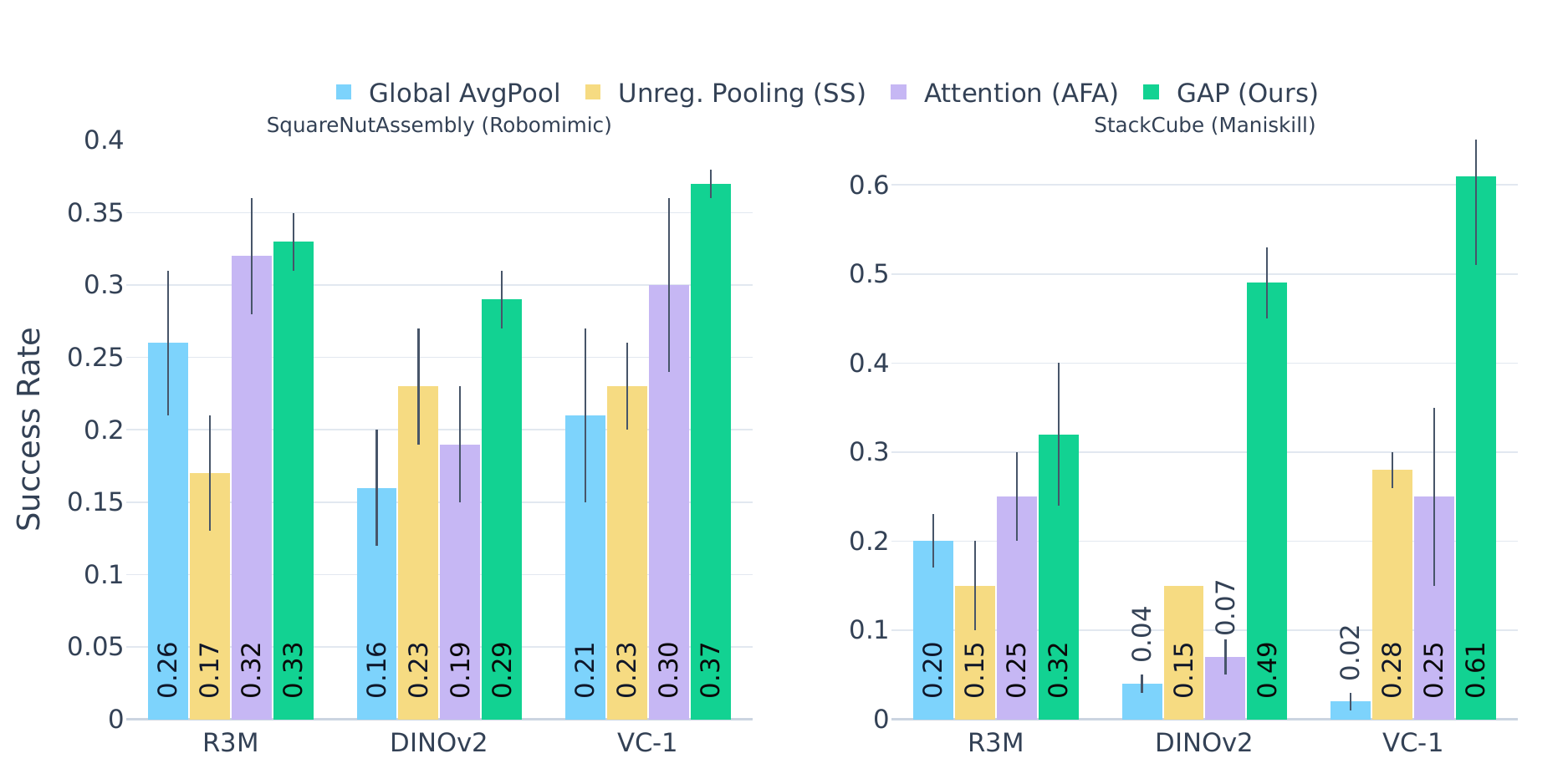}
    \caption{\textbf{GAP impact on different backbones (Square, 30 Demos).} We evaluate various pretrained backbones (R3M \cite{nair2023r3m}, VC1 \cite{majumdar2023vc1} and DinoV2 \cite{oquab2023dinov2}) with different poolers: a Global Avg. Pooling (blue), unregularized geometric adapter with Spatial Softmax (yellow), AFA \cite{tsagkas2025attentivefeatureaggregationor} (purple) and a GAP-pretrained spatial adapter (green). GAP consistently outperforms other methods by a large margin, demonstrating that all backbones benefit from our pretraining. %
    Results averaged over three seeds.
    }
    \label{fig:backbone_abl}
\end{figure*}

%% file: sections/loss_ablation_table.tex
\begin{table}[h]
\centering
\sffamily 
\renewcommand{\arraystretch}{1.1} 

\resizebox{0.8\columnwidth}{!}{%
\begin{tabular}{ccc cc}
\toprule
\textbf{$\mathcal{L}_{cen}$} & \textbf{$\mathcal{L}_{div}$} & \textbf{$\mathcal{L}_{spr}$} & \textbf{Success Rate} & \textbf{$\Delta$} \\
\midrule

\rowcolor{TechTeal} \cmark & \cmark & \cmark & \textbf{0.37} & \\

\addlinespace[0.2ex] 
\cmark & \cmark &        & 0.24 & \drop{0.13} \\
\cmark &        & \cmark & 0.32 & \drop{0.05} \\
       & \cmark & \cmark & 0.26 & \drop{0.11} \\

\cmark &        &        & 0.30 & \drop{0.07} \\
       &        & \cmark & 0.26 & \drop{0.11} \\
       & \cmark &        & 0.22 & \drop{0.15} \\

\bottomrule
\end{tabular}%
} %

\caption{\textbf{Loss Ablation (Square, 30 Demos).} Success rates when removing one or two loss components. Removing any individual term or combination degrades performance, proving that all the three terms synergistically contribute to the GAP objective.\label{table:loss_ablation}}

\end{table}

%% file: sections/pretr_ablation.tex
\begin{table}[h]
\centering
\sffamily 
\renewcommand{\arraystretch}{1.2} %

\begin{tabular}{l cc}
\toprule
\textbf{Pretraining Objective} & \textbf{Base: VC-1 + SS} & \textbf{Base: VC-1 + AFA} \\
\midrule

Baseline (No Pretraining) & \res{0.23}{0.03} & \res{0.31}{0.05} \\
+ Policy Training & \res{0.19}{0.04}\drop{0.04} & \res{0.27}{0.05}\drop{0.04} \\

\addlinespace[0.5ex] 
\rowcolor{TechTeal} \cellcolor{white}{} \textbf{+ GAP (Ours)} & \best{0.37}{0.06}\rise{0.14} & \na \\

\bottomrule
\end{tabular}
\caption{\textbf{Pretraining Impact (Square, 30 Demos).} Comparison of pretraining objectives added to standard baselines. Pretraining the encoder on the proxy task for policy learning actively degrades performance, whereas our GAP objective yields substantial improvements.}
\label{table:pretrain_abl}
\end{table}
\vspace{-5pt}

%% file: sections/5_conclusions.tex
\vspace{-7pt}
\section{Conclusion}
\label{sec:conclusion}
This paper identifies \emph{bottleneck collapse} as a key failure mode of frozen-VFM pipelines for few-shot visuomotor imitation learning: when only few demonstrations are available, the spatial pooling layer tends to overfit, losing object-centric geometric grounding and producing brittle representations.
Empirically, we observe that highly parameterized semantic poolers (e.g., AFA) can drift toward diffuse, unstable attention, while unregularized Spatial Softmax can lock onto arbitrary visual shortcuts.
To prevent this, we introduce \textbf{Geometric Anchor Pre-training (GAP)}, which regularizes the spatial bottleneck \emph{before} downstream policy learning using a geometric objective composed of centroid alignment ($\mathcal{L}_{center}$), geometric spread ($\mathcal{L}_{spread}$), and keypoint diversity ($\mathcal{L}_{div}$).
Pre-training the adapter once on a simple, cheap proxy task (RoboMimic \texttt{LiftCube}) produces transferable geometric anchors that can be reused across tasks and simulators.
Across four tasks, three backbone architectures, and three proxy visual domains, GAP consistently outperforms all baselines, improving both data efficiency and robustness under domain shift.

\textbf{Limitations.} GAP relies on ground-truth object segmentation masks during proxy pre-training, which typically requires simulator access.
In fully real-world settings without a simulator counterpart, masks would need to be obtained from an off-the-shelf segmentation model (e.g., Segment Anything), which may introduce noise and bias into the geometric supervision \cite{kirillov2023segment}.
In addition, while our proxy-domain ablations indicate robustness to simulator and table changes, we have not yet validated GAP on deformable objects or highly irregular geometries, where centroid-based supervision may be less informative.

\textbf{Future Work.} Our results suggest a clear complementarity between frozen VFMs and GAP: VFMs provide strong semantic representations, while GAP provides precise geometric grounding tied to object structure.
A natural extension is to explicitly fuse these signals by combining VFM semantic embeddings with GAP's coordinate-based spatial representation to obtain policies that are simultaneously semantically robust and geometrically precise.
A second direction would be study the effectiveness of GAP on text-conditioned models.
\vspace{-3pt}